\documentclass[11pt,a4paper]{article}
\usepackage[hyperref]{emnlp-ijcnlp-2019}
\usepackage{times}
\usepackage{latexsym}

\usepackage{url}
\usepackage{CJK}
\usepackage{multirow}
\usepackage{graphicx}
\usepackage{amsmath}
\usepackage{amssymb}
\newcommand\vecf[1]{\overrightarrow{#1}}
\newcommand\cevf[1]{\overleftarrow{#1}}
\usepackage{todonotes}
\usepackage{booktabs}

\aclfinalcopy 

\title{Semi-supervised Text Style Transfer: Cross Projection in Latent Space}

\author{Mingyue Shang\textsuperscript{1}\thanks{This work was done when Mingyue Shang was an intern at Tencent AI Lab.}, Piji Li\textsuperscript{2}, Zhenxin Fu\textsuperscript{1}, Lidong Bing\textsuperscript{4}, Dongyan Zhao\textsuperscript{1,3}, Shuming Shi\textsuperscript{2}, Rui Yan\textsuperscript{1,3}\thanks{Corresponding author: Rui Yan (ruiyan@pku.edu.cn)} \\
\textsuperscript{1}{Wangxuan Institute of Computer Technology, Peking University, China} \\
\textsuperscript{2}{Tencent AI Lab, Shenzhen, China}, \textsuperscript{3}{Center for Data Science, China} \\
\textsuperscript{4}{R\&D Center Singapore, Machine Intelligence Technology, Alibaba DAMO Academy} \\
\{shangmy, fuzhenxin, zhaodongyan, ruiyan\}@pku.edu.cn\\ \{pijili, shumingshi\}@tencent.com, l.bing@alibaba-inc.com
}

\date{}

\begin{document}
\maketitle
\begin{abstract}
Text style transfer task requires the model to transfer a sentence of one style to another style while retaining its original content meaning, which is a challenging problem that has long suffered from the shortage of parallel data. In this paper, we first propose a semi-supervised text style transfer model that combines the small-scale parallel data with the large-scale nonparallel data. With these two types of training data, we introduce a projection function between the latent space of different styles and design two constraints to train it. We also introduce two other simple but effective semi-supervised methods to compare with. To evaluate the performance of the proposed methods, we build and release a novel style transfer dataset that alters sentences between the style of ancient Chinese poem and the modern Chinese. 
\end{abstract}

\section{Introduction}

Recently, the natural language generation (NLG) tasks have been attracting the growing attention of researchers, including response generation \cite{vinyals2015neural}, machine translation \cite{bahdanau2014neural}, automatic summarization  \cite{chopra2016abstractive}, question generation \cite{Gao2018DifficultyCQ}, etc. Among these generation tasks, one interesting but challenging problem is text style transfer \cite{shen2017style, fu2018style, logeswaran2018content}. Given a sentence from one style domain, a style transfer system is required to convert it to another style domain as well as keeping its content meaning unchanged. As a fundamental attribute of text, style can have a broad and ambiguous scope, such as ancient poetry style v.s. modern language style and positive sentiment v.s. negative sentiment.

Building such a style transfer model has long suffered from the shortage of parallel training data, since constructing the parallel corpus that could align the content meaning of different styles is costly and laborious, which makes it difficult to train in a supervised way. Even some parallel corpora are built, they are still in a deficient scale for neural network based models. To tackle this issue, previous works utilize nonparallel data to train the model in an unsupervised way. 
One commonly used method is disentangling the style and content from the source sentence \cite{john2018disentangled, shen2017style, hu2017toward}. For the input, they learn the representations of styles and style-independent content, expecting the later only keeps the content information. Then the content-only representation is coupled with the representation of a style that differs from the input to produce a style-transferred sentence. The crucial part of such methods is that the encoder should accurately disentangle the style and content information. However, \citet{lample2018multipleattribute} illustrated that disentanglement is not a facile thing and the existing methods are not adequate to learn the style-independent representations.


Considering the above-discussed issues, in this paper, instead of disentangling the input, we propose a differentiable encoder-decoder based model that contains a projection layer to build a bridge between the latent spaces of different styles. Concretely, for texts in different styles, the encoder converts them into latent representations in different latent spaces. We introduce a projection function between two latent spaces that projects the representation in the latent space of one style to another style. Then the decoder generates an output using the projected representation.

In the majority of cases, nonparallel corpora of different styles are accessible. Based on these datasets, it is feasible to build small-scale parallel datasets. Therefore, we design two kinds of objective functions for our model so that it could be trained under both supervised settings and unsupervised settings. With the parallel data, the model learns the projection relationship and the standard representation of the target latent space from ground-truth. Without the parallel data, we train the model by back-projection between the source and target latent space so that it could learn from itself. During training, we incorporate these two kinds of signals to train the model in a semi-supervised way.

We conduct experiments on an English formality transfer task that alters the sentence between formal and informal styles, and a Chinese literal style transfer task that alters between the ancient Chinese poem sentences and modern Chinese sentences. We evaluate the performance of models from the degree of content preservation, the accuracy of the transferred styles, and the fluency of sentences using both automatic metrics and human annotations. Experimental results show that our proposed semi-supervised method can generate more preferred output.

In summary, our contributions are manifolds:
\vspace{-3mm}
\begin{itemize}
\setlength{\itemsep}{0pt}
\setlength{\parsep}{0pt}
\setlength{\parskip}{0pt}
\item We designed a semi-supervised model that crossly projects the latent spaces of two styles to each other. In addition, this model is flexible in alternating the training mode between supervised and unsupervised. 
\item We introduce another two semi-supervised methods that are simple but effective to leverage both the nonparallel and parallel data.
\item We build a small-scale parallel dataset that contains ancient Chinese poem style and modern Chinese style sentences. We also collect two large nonparallel datasets of these styles.\footnote{Download link: https://tinyurl.com/yyc8zkqg} 
\end{itemize}

\section{Related Works}
Recently, text style transfer has stimulated great interests of researchers from the area of neural language processing and some encouraging results are obtained \cite{shen2017style,N18-1012,P18-1080,hu2017toward,jin2019unsupervised}.
In the primary stage, due to the lacking of parallel corpus, most of the methods employ unsupervised learning paradigm to conduct the semantic modeling and transfer.

\paragraph{Unsupervised Learning Methods.} 
\citet{mueller2017sequence} modify the latent variables of sentences in a certain direction guided by a classifier to generate the sentence with classifier favored style.
\citet{hu2017toward} employ variational auto-encoders \cite{DBLP:journals/corr/KingmaW13} to conduct the style latent variable learning and design a strategy to disentangle the variables for content and style.
\citet{shen2017style} first map the text corpora belonging to different styles to their own space respectively, and leverages the alignment of latent representations from different styles to perform style transfer.
\citet{chen2019unsupervised} propose to extract and control the style of the image caption through domain layer normalization.
\citet{P18-1080} and \citet{zhang2018style} employ the back-translation mechanism to ensure that the input from the source style can be reconstructed from the transferred result in the target style.
\citet{liao2018quase} associate the style latent variable with a numeric discrete or continues numeric value and can generate sentences controlled by this value. 
Among them, many use the adversarial training mechanism \cite{goodfellow2014generative} to improve the performance of the basic models \cite{shen2017style,zhao2018language}. 

To sum up, most of the existing unsupervised frameworks on the text style transfer focus on getting the disentangled representations of style and content. 
However, \citet{lample2018multipleattribute} illustrated that the disentanglement not adequate to learn the style-independent representations, thus the quality of the transferred text is not guaranteed.

\paragraph{Supervised Learning Methods.}
Fortunately, \citet{N18-1012} release a high-quality parallel dataset - GYAFC, which consists of two domains for Formality style transfer: Entertainment \& Music and Family \& Relationships. The quantity of the dataset is sufficient to train an attention-based sequence-to-sequence framework.

Nevertheless, building parallel corpus that could align the content meaning of different styles is costly and laborious, and it is impossible to build the parallel corpus for all the domains.

\section{Task Formulation}
We assume two nonparallel datasets $A$ and $B$ that are sentences in two different styles, denoted as style $s^a$ and style $s^b$, and a parallel dataset $P$ that contains the pairs of sentences with two styles. The size of the three datasets are denoted as $|A|$, $|B|$ and $|P|$, respectively. As the nonparallel data is abundant but the parallel data is limited, size $|A|$, $|B| \gg |P|$. Formally, two nonparallel datasets are denoted as $A = \{a_1, a_2, \cdots, a_{|A|}\}$ and $B = \{b_1, b_2, \cdots, b_{|B|}\}$, where ${a_i}$ and ${b_i}$ refer to the $i$-th sentences of $s^a$ style and $s^b$ style, respectively. It should be clear that as $A$ and $B$ are nonparallel datasets, the sentences in the two datasets are not aligned with each other, which means that $a_i$ and $b_i$ are not required to have the same or similar content meaning though they have the same subscript. The parallel dataset is denoted as $P = \{(a^p_1, b^p_1), (a^p_2, b^p_2), \cdots, (a^p_{|P|}, b^p_{|P|})\}$, where $(a^p_i, b^p_i)$ is the $i$-th pair of sentences that have the same content meaning but expressed in style $s^a$ and $s^b$ separately.


Due to the situation that the limited parallel data is deficient for neural network based model to train, our goal is to train a model in a semi-supervised way that could leverage the large volumes of nonparallel data to improve the performance. Given a sentence of the source style as input, the model learns to generate a target style output which preserves the meaning of the input to the greatest extent. In this paper, the style transfer process could be exerted in two directions, from $s^a$ to $s^b$ as well as from $s^b$ to $s^a$.

\begin{figure*}
    \centering
    \includegraphics[width=15cm]{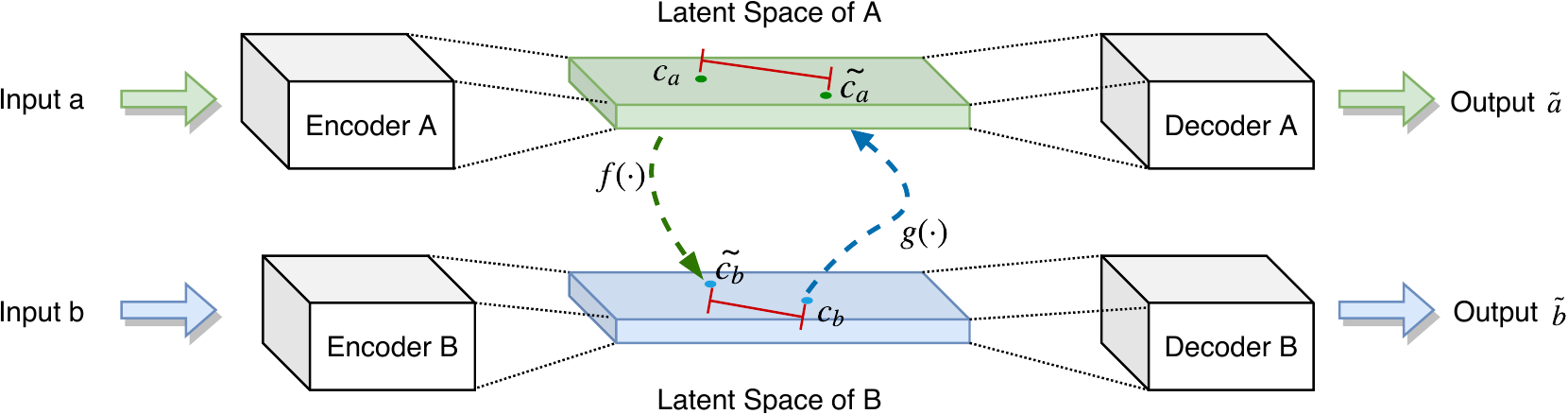}
    \caption{The process of training the sequence-to-sequence model. Take a sentence pair $(a, b)$ for example. After $a$ is encoded into $c_a$ by the encoder A, the projection function $f(\cdot)$ (shown by the dotted green line) projects it to the latent space of $B$, which is then fed to the decoder B to generate the output $\tilde{a}$. The encoder B gives the latent representation of $b$. One of the constraints is computed from the distance of $\tilde{c}_b$ and $c_b$ (shown by the red lines) and another is the NLL-loss of $\tilde{b}$ and $b$. The dotted blue line refers to the projection function from the space of $B$ to $A$.}
    \label{fig:model}
\end{figure*}

\section{Basic Model Architecture}
The text style transfer task could be interpreted as transforming a sequence of source style words to a sequence of target style words, which makes the sequence-to-sequence framework a suitable architecture. In this section, we describe the formulation of sequence-to-sequence (S2S)~\cite{sutskever2014sequence} model, which contains an encoder and a decoder, and our proposed models are built based on such an architecture. 

Given an input, the encoder first converts it into an intermediate vector, then the decoder takes the intermediate representation as input to generate a target output. In this paper, we implement the encoder by a bi-directional Long Short-Term Memory (BiLSTM) \cite{hochreiter1997long} and the decoder by a one layer LSTM. 

Formally, with an input sentence $x =  \{w^x_1, \cdots, w^x_T\}$ of length $T$, and a target sentence $y = \{w^y_1, \cdots, w^y_{T'}\}$ of length $T'$, where $w_i$ is the embedding of the $i$-th word, the probability of generating the target sentence $y$ given $x$ is defined as:
\begin{equation}
    p(y|x) = \prod_{i=1}^{T'}p\left( w^y_i | x, w^y_1, \ldots , w^y_{i-1} \right)
    \label{prob}
\end{equation}

In the training process, the encoder first encodes $x$ into an intermediate representation $c_x$. More specifically, at each time step $t$, the encoder produces a hidden state vector $h_t = [\vecf{h}_t; \cevf{h}_t]$, where the $\vecf{h}_t$ and $\cevf{h}_t$ represents the forward and backward hidden state, respectively, and $[;]$ means concatenation. 
The intermediate vector is formed by the concatenation of the forward and backward last hidden states, denoted as $c_x = Enc(x) = [\vecf{h}_T; \cevf{h}_1]$. Then this vector is fed to the decoder to generate a target sentence step by step as formulated in Equation \ref{prob}.


Such neural network based models always have a large amount of parameters which empower the model with the ability to fit complex datasets. However, when trained with limited data, the model may get so closely fitted to the training data that loses the power of generalization, and thus performances poorly on new data.


\section{Cross Projection in Latent Space}
In this paper, we propose a semi-supervised style transfer method named cross projection in latent space (CPLS) to leverage the large volumes of nonparallel data as well as the limited parallel data. Based on the S2S architecture, we consider the representations in the latent space, and propose to establish projection relations between different latent spaces. To be specific, in the S2S architecture, the encoder converting the input into an intermediate vector is a process of extracting the semantic information of the input, including the style and the content information. Thus the intermediate vectors could be seen as the compressed representations in the latent space of the inputs. Since texts in different styles are in different latent spaces, we introduce a projection function to project the intermediate vector from one latent space to another. To combine the nonparallel data and parallel data in training, we design two kinds of constraints for the projection functions. 

Concretely, we first train an auto-encoder for each style to learn the standard latent spaces of the style.
After that, we train the projection functions which are exerted on latent vectors to establish projection relations between different latent spaces. We design a cross projection strategy and a cycle projection strategy to utilize the parallel data and nonparallel data. These two strategies are exerted iteratively in the training process, thus our model is compatible with the two types of training data. The following subsections give the details of the modules and the strategies of model training.


\subsection{Denoising Auto-Encoder}
To learn the latent space representations of styles, we train an auto-encoder for each style of text through reconstructing the input, which is composed of an encoder and a decoder. The encoder takes a sentence as input and maps the sentence to a latent vector representation, and the decoder reconstructs the sentence based on the vector. But a common pitfall is that the decoder may simply copy every token from the input, making the encoder lose the ability to extract the features.  To alleviate this issue, we adopt the denoising auto-encoder (DAE) following the previous work \cite{lample2018multipleattribute} that tries to reconstruct the corrupted input. Specifically, we randomly shuffle a small portion of words in the original sentence to the corrupted input which are then fed to the denoising auto-encoder. For example, the original input is ``Listen to your heart and your mind.'', then the corrupted version could be ``Listen your to and heart your mind.''. The decoder is required to reconstruct the original sentence. 

The training of a DAE model relies on the corpus of one style only, thus we train each DAE model using the nonparallel corpus of each style. Formally, the encoder and decoder of style $s^a$ and $s^b$ are referred as $\text{Enc}^{(a)}$ - $ \text{Dec}^{(a)}$ and $\text{Enc}^{(b)}$ - $\text{Dec}^{(b)}$. 
Given sentences $a$ and $b$ from two styles, the corresponding encoder takes the sentence as input and converts it to the latent vector $c_a$ and $c_b$, respectively, by which each encoder therefore constructs the latent space for the corresponding style.

\subsection{Cross Projection}
To perform the style transfer task, we establish a projection relationship between latent spaces and cross link the encoder and the decoder from different styles. 
Take the transfer from style $s^a$ to $s^b$ for example. Given a sentence $a$ that is required to transferred to $b$, we cross link the $\text{Enc}^{(a)}$ as the encoder and $\text{Dec}^{(b)}$ as the decoder to form the style transfer model. However,
considering the DAE models for different styles are trained respectively, therefor the latent vector spaces are usually different.
The latent vector produced by $\text{Enc}^{(a)}$ is the representation on the latent space of style $s^a$ while the $\text{Dec}^{(b)}$ relies on the information and features from the latent space of style $s^b$. Therefore, we introduce a projection function to project the latent vector from the latent space of $s^a$ to $s^b$.

Concretely, after we get the $c_a$ from $\text{Enc}^{(a)}$, a projection function $f(\cdot)$ is employed to project $c_a$ from the latent space of style $s^a$ to the latent space of $s^b$, denoted as $\tilde{c}_b = f(c_a)$. Then the decoder $\text{Dec}^{(b)}$ takes the the projected vector $\tilde{c}_b$ as input and generates a sentence $\tilde{b}$ of style $s^b$ base on the prediction probability, denoted as 
$p_b(\tilde{b}|\tilde{c}_b)$.

It is worth noting that we have only exploited the nonparallel corpus for the style transfer up to now.
Recall that our framework can employ both the parallel corpus and nonparallel corpus for the model training. With parallel corpus, we design the cross projection strategy. When the input $a$ is accompanied with a ground-truth $b$, we can get the standard latent representation of $b$ by the $\text{Enc}^{(b)}$, denoted as $c_b$. In order to align the latent vectors from different spaces, we design the constraints to train the $f(\cdot)$ from two aspects: the distance between $\tilde{c}_b$ and $c_b$ in the latent space should be close; the generated sentence $\tilde{b}$ should be similar with $b$. Then we define two losses as follows:
\begin{align}
    \tilde{c}_b =~&f(\text{Enc}^{(a)}(a)) \\
    l_{s_1} =~&\|\tilde{c}_b - c_b\| \\
    l_{s_2} =~&- \log p(b|\tilde{c}_b) \\
    l_s =~\alpha &*  l_{s_1} + \beta * l_{s_2} 
\end{align}
where $l_{s_1}$ measures the Euclidean distance between $\tilde{c}_b$ and $c_b$ and $l_{s_2}$ is the negative log-likelihood (NLL) loss given $a$ as input and $b$ as the ground-truth. $\alpha$ and $\beta$ is the hyper-parameters that control the weight of $l_{s_1}$ and $l_{s_2}$. 

Figure \ref{fig:model} shows the training process of the cross projection. Similarly, to transfer the style of text from $s^b$ to $s^a$, we cross link $\text{Enc}^{(b)}$ and $\text{Dec}^{(a)}$, and the projection function to project $c_b$ to the latent space of $s^a$ is denoted as $g(\cdot)$.

\begin{figure}
    \centering
    \includegraphics[width=5cm]{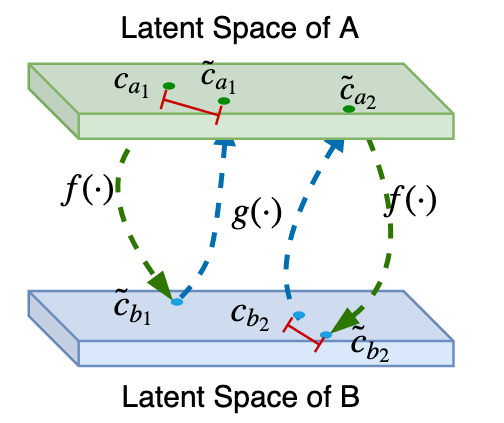}
    \caption{The cycle projection process between the latent space of $A$ and $B$ when training the denoising auto-encoders. The encoders and decoders are the same as illustrated in Figure \ref{fig:model} and are omitted in this figure.}
    \label{fig:cycle}
\end{figure}

\subsection{Cycle Projection Enhancement}
Inspired by the concept of back-translation in machine translation \citep{sennrich2015improving, lample2018phrase, he2016dual}, we design a cycle projection strategy to train the model on the nonparallel data.
Given the input without the ground-truth, we train the projection function $f(\cdot)$ and $g(\cdot)$ by projecting back and forth to reconstruct the input. The cycle projection process is shown in Figure \ref{fig:cycle}.

Formally, for a sentence $a$ in style $s^a$, after getting its latent representation ${c_a}$ in its own latent space by the $\text{Enc}^{(a)}$, we first project it to the latent space of $s^b$ by $f(\cdot)$ and get $\tilde{c}_b = f(c_a)$. Then we exert $g(\cdot)$ to project $\tilde{c}_b$ back to the latent space of style $s^a$, denoted as $\tilde{c}_a = g(\tilde{c}_b)$. Finally $\tilde{c}_a$ is fed to $\text{Dec}^{(a)}$ to produce an output $\tilde{a}$. Though the latent vector $\tilde{c}_b$ has no reference to train, the latent vector $\tilde{c}_a$ and the output $\tilde{a}$ could be trained by treating $c_a$ and $a$ as the references. The loss functions are formulated as:
\begin{align}
    \tilde{c}_a =~&g(f(\text{Enc}^{(a)}(a))) \\
    l_{c_1} =~&\|\tilde{c}_a - c_a\| \\
    l_{c_2} =~&- \log p(a|\tilde{c}_a) \\
    l_c = \alpha &*  l_{c_1} + \beta * l_{c_2} 
\end{align}


\subsection{Training Procedure}
In the training stage, we first pretrain DAE models for each style separately on the nonparallel data to get the latent spaces of styles. Then we alternately apply the cross projection strategy and cycle projection enhancement on parallel data and nonparallel data with loss $l_s$ and $l_c$. 

\section{Straightforward Semi-supervised Methods}
\label{semi-base}
We also introduce another two semi-supervised methods as the baselines to provide a more comprehensive comparison with the proposed CPLS model. The two semi-supervised methods are built from different perspectives.

\subsection{Data Augmentation via Retrieval}
One semi-supervised baseline is from the perspective of data augmentation. In order to alleviate the over-fitting issue caused by the small-scale parallel data, we propose a simple but efficient method that augments the parallel dataset by retrieving pseudo-references for the nonparallel datasets, denoted as DAR model. 

\paragraph{Pseudo-parallel Corpus Construction.} 
We employ Lucene\footnote{\url{http://lucene.apache.org/}} to build indexes for the nonparallel corpus of each style. Then the pseudo-references are retrieved based on the TF-IDF scores. To build the pseudo-parallel corpus of style $s^a$ and $s^b$, we 
first samples 80,000 sentences in style $s^a$ as queries. Specifically, each query searches a pseudo-reference from the nonparallel corpus of style $s^b$ according to the TF-IDF based cosine similarity. The sampled query is therefore coupled with the pseudo-reference to form a training pair. We also conduct the same operation that uses sampled queries from style $s^b$ to search pseudo-references from sentences of style $s^a$. 
After searching pseudo-references from two sides, we construct a pseudo-parallel corpus in the size of 150,000 pairs\footnote{Some quires did not have corresponding retrieved result, therefore the total size is less than the number of queries.}. With the pseudo-parallel corpus, the model is exposed to more information and thus the problem of over-fitting can be mitigated to some extent. Though the relevance of the content between the input sentence and the pseudo-reference is not guaranteed, the encoder could better learn to extract the language features and the decoder could also benefit from the weak contextual correlation information.

\paragraph{Training Procedure.}  In the training stage, we first train the S2S model on the pseudo-parallel dataset and save a checkpoint every 2,000 steps. We then calculate the BLEU score of checkpoints on the validation set and select the one with the highest BLEU score as the final pretrained model. Then based on this pretrained model, we fine-tune the parameters using the true parallel data.
\vspace{-2mm}
\subsection{Shared Latent Space Model}
The second semi-supervised baseline is similar to CPLS that first trains a DAE model on the corpus of each style. Instead of building a bridge between the latent spaces of two styles through projection functions, in this method, we simply share the latent representations by cross linking the encoder and decoder, denoted as SLS model.
Given a pair of sentence $(a^p, b^p)$, the encoder $\text{Enc}^{(a)}$ encodes it into context $c_a$, then the decoder $\text{Dec}^{(b)}$ directly takes $c_a$ to produce $\tilde{b}$.

\paragraph{Training Procedure.} For this method, we first pre-train two DAE models. Then we train the DAE models and the cross-linked S2S models from two transfer directions alternately. Considering the size imbalance between the nonparallel corpora and parallel corpus, to avoid the S2S falling into over-fitting too fast, We alternate the training in the form of training 20-step DAE models and then training one step S2S model.

\begin{table}[]
    \centering
    \scalebox{0.9}{
    \begin{tabular}{c|c|c|c|c}
    \toprule[1pt]
        \multirow{2}*{Dataset}  &  \multicolumn{3}{c|}{Parallel} & \multirow{2}*{Nonparallel}  \\ \cline{2-4}
        & Train & Valid & Test    &  \\ \hline 
        Anc.p& \multirow{2}*{3,362} & \multirow{2}*{400}  & \multirow{2}*{998}  & 269k   \\
        M.zh &  &   &   & 77k    \\ \hline 
        F.en    & \multirow{2}*{5,000} & 2,247 & 1,019 & 2,483k \\ 
        Inf.en  &  & 2,788 & 1,332 & 4,891k \\ \bottomrule[1pt]
    \end{tabular}}
    \caption{Statistics of two style transfer datasets.}
    \label{tab:dataset}
\end{table}

\section{Experiment}
We conduct experiments on two bilateral style transfer tasks that each of them has a small-scale parallel corpus and two large-scale nonparallel corpora in two styles. 
DAR model and SLS model are treated as two baselines of CPLS model. In addition, we also train a S2S model with attention mechanism on the parallel data as the vanilla baseline. The following subsections elaborate the construction of the datasets and the detailed experimental settings.
\begin{CJK}{UTF8}{gkai}
\begin{table*}[]
    \centering
    \small
    \scalebox{0.95}{
    \begin{tabular}{c|c|l}
    \toprule[1pt] \hline
    \multirow{10}*{\rotatebox[origin=c]{90}{to Anc.p}} & \multirow{2}*{Source} & 问客人为什么来，客人说为了上山砍伐树木来买斧头。\\
         & & (Ask the guests why, the guest said he want to buy an axe in order to cut  the trees at the mountain.) \\ 
         & \multirow{2}*{S2S} & 客问谁客中，树。 \\ 
         & & (The guest asks who is in the guest, the tree) \\ 
         & \multirow{2}*{SLS} & 问人何为人， \\  
         & & (Ask people what people are,) \\ 
         & \multirow{2}*{DAR} & 客中何为客，山头为木头。 \\ 
         & & (What is the guest in the guest, the mountain head is the wooden head.) \\ 
         & \multirow{2}*{CPLS} & 问客何为焉，客从满木盘。 \\
         & & (Ask the guest what he is going for, the guest need to get something out of full wooden forest.) \\  \hline
          \multirow{10}*{\rotatebox[origin=c]{90}{to M.zh}} & \multirow{2}*{Source} & 水寒风似刀。 \\  
        & & (The water is as cold as a knife.)\\ 
         & \multirow{2}*{S2S} & 水夜寒风吹动了寒冷。 \\
         & & (The cold wind with water at night blew the cold )\\ 
         & \multirow{2}*{SLS} & 寒月寒风吹得寒冷寒冷。 \\ 
         & & (The cold wind and the cold moon blew the cold )\\ 
         & \multirow{2}*{DAR} & 寒风吹得很寒冷。 \\  
         & & (The cold wind is very cold)\\ 
         & \multirow{2}*{CPLS} & 水边寒冷的寒风吹过好像寒冷的刀刀。 \\  
         & & (The cold wind from the water side like a cold knife.)\\ \hline
        \multirow{5}{*}{\rotatebox[origin=c]{90}{to F.en}} & Source & give them a chance to discover you. \\ 
        & S2S & Share them an opportunity to meet you. \\ 
        & SLS & I give them a chance to discover you. \\  
        & DAR & Give them a chance to discover you. \\  
        & CPLS & You should give them a chance to discover you. \\ \hline
        \multirow{5}*{\rotatebox[origin=c]{90}{to Inf.en}} & Source & I think it is wrong that they cannot go out with her. \\   
        & S2S & I think it is wrong that is wrong with her. \\ 
        & SLS & I think they can't go out with her. \\ 
        & DAR & I think it is wrong that they cant go out with her. \\ 
        & CPLS & It is wrong that they can't go out with her. \\  \hline \bottomrule[1pt]
    \end{tabular}}
    \caption{Case study: the sentences in the brackets are the translation of the corresponding sentence.}
    \label{cases}
\end{table*}
\end{CJK}

\subsection{Datasets}
We construct a Chinese literal style dataset with ancient poetry style and modern Chinese style, and an English formality style dataset with formal style and informal style. The texts of ancient Chinese poem, modern Chinese, formal English and informal English are referred as Anc.P, M.zh, F.en and Inf.en for short. The overall statistics are shown in Table \ref{tab:dataset}.

\noindent \textbf{Chinese Literal Style Dataset.}
We consider the texts of ancient Chinese poem style and modern Chinese style which differ greatly in expression. As the ancient poems from different dynasties vary in style and form, in this paper, we focus on the poem from Tang dynasty. 

To build the parallel dataset, we crawl from the website named Gushiwen\footnote{\url{https://www.gushiwen.org/}} that provides ancient Chinese poems and some are coupled with modern Chinese interpretation in paragraphs. 
We split the collected paragraphs into independent sentences using punctuation based rules, and manually align the poem sentence with the interpretation sentence to form the parallel pair.

For the nonparallel corpus of ancient Chinese poem style, we collect all the poems from Quan Tangshi\footnote{The largest collection of Tang poems commissioned by the Qing dynasty Kangxi Emperor.} and split them into sentences. To build the nonparallel dataset in modern Chinese style, we collect data from the Chinese ballad and took the lyrics. The reason we choose the Chinese ballad is that the content domain of the two styles should be close. Considering that most of the poems are about the natural scenery and the sentiments, the most suitable literary form is the lyrics of ballad. \footnote{ We count the overlap of the topic words involved in the top 20 topics between the poem translation and ballad, and compare the overlap ratio with other modern Chinese corpora (news, open-domain conversations).}



\noindent \textbf{Formality Dataset.}
The formality dataset used in this paper is built based on the parallel dataset released by \citet{N18-1012} which contains texts of formal and informal style. 

With the released data, we randomly sample 5,000 sentence pairs from it as the parallel corpus with limited data volume. 
We then use the Yahoo Answers L6 corpus\footnote{\url{https://webscope.sandbox.yahoo.com/catalog.php?datatype=1}} as the source which is in the same content domain as the parallel data to construct the large-scale nonparallel data. To divide nonparallel dataset into two styles, we train a CNN-based classifier~\citep{kim2014convolutional} on the parallel data with annotation of styles and use it to classify the nonparallel data. 

\begin{table*}[!tp]
    \centering
    \scalebox{0.84}{
    \begin{tabular}{c|c|c|c|c|c|c|c|c|c|c|c|c}
    \toprule[1pt]
    \hline
          & \multicolumn{3}{c|}{to Anc.P} & \multicolumn{3}{c|}{to M.zh} & \multicolumn{3}{c|}{to F.en} & \multicolumn{3}{c}{to Inf.en} \\ \hline
        Model    & Acc & BLEU & GLEU & Acc & BLEU & GLEU & Acc & BLEU & GLEU & Acc & BLEU & GLEU  \\ \hline
        S2S      & \bf 87.2\% & 4.20     & 3.24 & 74.8\%     & 3.66     & 3.43 & 88.9\%         & 33.90     & 14.06 & \bf 71.8\% & 18.34 & 2.99 \\
        SLS      & 82.0\%     & 5.89     & 4.49 & 81.9\%     & 3.05     & 1.88 & 89.5\%         & 41.41     & 16.77 & 63.5\%     & 19.21 & 2.55 \\
        DAR      & 82.5\%     & 6.33     & 5.21 & 80.4\%     & 4.72     & \bf 4.26 & 89.2\%     & 44.72     & 18.52 & 63.5\%     & 23.32 & 3.26 \\
        CPLS     & 85.4\%     & \bf 7.11 & \bf 5.52 & \bf 84.4\% & \bf 4.95 & 4.12 & \bf 91.3\% & \bf 48.60 & \bf 19.04 & 64.3\%     & \bf 27.25 & \bf 3.61 \\ \hline \bottomrule[1pt]
    \end{tabular}}
    \caption{Automatic evaluation results on four style transfer tasks. Acc refers to the style accuracy.}
    \label{auto results}
\end{table*}

\begin{table}[]
    \centering
    \small
    \begin{tabular}{c|l|c|c|c}
    \toprule[1pt]
        Model & Dataset & Content & Style & Fluency  \\ \hline
        \multirow{4}*{S2S} & to M.zh& 0.1875 & 0.5675 & 0.3575\\ 
        & to Anc.P      & 0.2275 & 0.5400 & 0.4425  \\
        & to Inf.en   & 0.3175 & 0.4600 & 0.5800  \\
        & to F.en     & 0.3625 & 0.5125 & 0.6325  \\   \hline
        \multirow{4}*{CPLS}& to M.zh & 0.3100 & 0.5825 & 0.3050\\
        & to Anc.P      & 0.4375 & 0.6875 & 0.5475 \\
        & to Inf.en   & 0.4675 & 0.4625 & 0.5725  \\
        & to F.en      & 0.4600 & 0.5675 & 0.6200  \\   \bottomrule[1pt]
    \end{tabular}
    \caption{The human annotation results of the S2S model and CPLS model from three aspects.}
    \label{results}
\end{table}

\subsection{Experimental Settings}
We perform different data preprocessing on different datasets. 
The Chinese literary datasets are segmented by characters instead of word to alleviate the issue of unknown words. Our statistics show that the average length of ancient poems is 9 while 17 for modern Chinese sentences. Therefore, we set the minimum length of the ancient poem as 3, and the maximum length of the modern Chinese sentence as 30. For the formality datasets, we use NLTK~\citep{loper2002nltk} to tokenize the texts and set the minimum length as 5 and the maximum length as 30 for both formal and informal styles.

We adopt GloVE \cite{pennington2014glove} to pretrain the embeddings, 
and the dimensions of the embeddings are set to 300 for all the datasets. The hidden states are set to 500 for both encoders and decoders. We adopt SGD optimizer with the learning rate as 1 for DAE models and 0.1 for S2S models. The dropout rate is 0.4. In the inference stage, the beam size is set to 5.

\section{Comparison Models and Evaluations}
\subsection{Baselines}
We train a S2S model with attention mechanism on the parallel data as the supervised learning baseline. Since the existing works on text style transfer seldom explore the semi-supervised methods, we propose DAR and SLS model as two semi-supervised baselines.

\subsection{Automatic Evaluation Metrics}
Following previous works \cite{P18-1080,D18-1138}, we employ BLEU score \cite{papineni2002bleu} and style accuracy as the automatic evaluation metrics to measure the content preservation degree and the style changing degree. BLEU calculates the N-gram overlap between the generated sentence and the references, thus can be used to measure the preservation of text content. Considering that text style transfer is a monolingual text generation task, we also use GLEU, a generalized BLEU proposed by \citep{napoles2015ground}. To evaluate the extent to which the sentences are transferred to the target style, we follow \citet{shen2017style,hu2017toward} that build a CNN-based style classifier and use it to measure the style accuracy. 

\subsection{Human Evaluation}
We also adopt human evaluations to judge the quality of the transferred sentences from three aspects, namely content, style and fluency. These aspects evaluate how well the transferred text preserve the content of the input, the style strength and the fluency of the transferred text. Take the content relevance for example, the criterion is as follows:~\textbf{+2:} The transferred sentence has the same meaning with the input sentence.~\textbf{+1:} The transferred sentence preserves part of the content meaning of the input sentence.~\textbf{0: } The transferred sentence and the input sentence are irrelevant in content. The criteria for style strength and fluency are similar to the content relevance criterion.

 To get the evaluation results, we first randomly sample 50 test cases from each dataset. As the style transfer is bilateral in this paper, there are 400 test cases in all. We invited four well-educated volunteers to score the results from the supervised baseline and the three semi-supervised models. 

\section{Results and Analysis}
\noindent \textbf{Evaluation Results.}
Table \ref{auto results} presents the evaluation results of automatic metrics on the models.
It can be seen that the BLEU scores and GLEU scores of the semi-supervised models on almost all the datasets are better than the baseline S2S model. This result indicates that the model benefits from the nonparallel data in terms of content preservation. One interesting thing is that the overall BLEU scores on the ancient poems and modern Chinese datasets are lower than other datasets. This result may be explained by the fact that the edit distance between formal and informal texts are smaller than between ancient poems and modern Chinese texts. Therefore, it is more challenging for model to preserve the content meaning when transferring between ancient poems and modern Chinese text. Among three semi-supervised models, CPLS model achieves the greatest improvement, verifying the effectiveness of the projection functions. 
However, the gain of CPLS model in the aspect of style accuracy is not that significant. A possible explanation may be the bias of the style classifier. Take the transfer task from ancient poems to modern Chinese text for example. We observe that the classifier tends to classify short sentences into ancient poems as length is an obvious feature. We analyse the sentences generated by S2S model and by the CPLS model, and the statistics show that the average length of the text generated by S2S model is shorter, which may lead to the bias of the style classifier. Therefore, we also adopt human evaluation to alleviate this issue.

Table \ref{results} compares the human evaluation results of S2S model and CPLS model on all the datasets, which are calculated by the average score of the human annotations. As shown in the Table \ref{results}, the CPLS model outperforms the S2S model in the aspects of the content preservation and style strength, and is on par in terms of fluency. 

\noindent \textbf{Case Study.}
We select the generated results of the S2S baseline and three semi-supervised models on two style transfer tasks due to the limited space as shown in Table \ref{cases}. Compared with the Chinese literacy style datasets, the formality datasets are less challenging as discussed before. Thus it can be seen from the table that the generated sentences on the formality datasets are more fluent. For the task that transferring from the modern Chinese text to the ancient poem, the S2S generates a shorter sentence while the CPLS model generates the sentence that preserve the most content information.

\section{Conclusion}
In this paper, we design a differentiable semi-supervised model that introduces a projection function between the latent spaces of different styles. The model is flexible in alternating the training mode between supervised and unsupervised learning. We also introduce another two semi-supervised methods that are simple but effective to use the nonparallel and parallel data. We evaluate our models on two datasets that have small-scale parallel data and large-scale nonparallel data, and verify the effectiveness of the model on both automatic metrics and human evaluation.

\section{Acknowledgement}
We would like to thank the anonymous reviewers for their constructive comments. This work was supported by the National Key Research and Development Program of China (No. 2017YFC0804001), the National Science Foundation of China (NSFC No. 61876196 and NSFC No. 61672058). Rui Yan was sponsored by Tencent Collaborative Research Fund.

\bibliography{emnlp-ijcnlp-2019}
\bibliographystyle{emnlp_natbib}

\end{document}